%% file: main.tex
\documentclass[10pt,twocolumn,letterpaper]{article}

\usepackage[pagenumbers]{iccv}


\definecolor{iccvblue}{rgb}{0.21,0.49,0.74}
\usepackage[pagebackref,breaklinks,colorlinks,allcolors=iccvblue]{hyperref}

\def\confName{ICCV}
\def\confYear{2025}

\title{EC-Flow: Enabling Versatile Robotic Manipulation from Action-Unlabeled Videos via \underline{E}mbodiment-\underline{C}entric Flow
}

\author{
Yixiang Chen$^{1,2}$ \ \
Peiyan Li$^{1,2}$ \ \
Yan Huang$^{1,2,3\dag}$ \ \
Jiabing Yang$^{1,2,3}$ \ \
Kehan Chen$^{1,2}$ \ \
Liang Wang$^{1,2\dag}$ \\
$^{1}$New Laboratory of Pattern Recognition (NLPR), \\Institute of Automation, Chinese Academy of Sciences \\ 
$^{2}$School of Artificial Intelligence, University of
Chinese Academy of Sciences ~~$^{3}$FiveAges\\
{\tt\small yixiang.chen@cripac.ia.ac.cn,  \{yhuang, wangliang\}@nlpr.ia.ac.cn}
}

\usepackage{amsfonts}
\usepackage{amsmath}
\usepackage{breqn}
\usepackage{booktabs}
\usepackage{graphicx}
\usepackage{amsmath}
\usepackage{adjustbox}
\usepackage{multirow}
\usepackage{bbding}
\usepackage{fontawesome5}
\usepackage[accsupp]{axessibility}  

\usepackage{algorithm}
\usepackage[noend]{algpseudocode}
\algrenewcommand{\algorithmicrequire}{\textbf{Input:}}
\algrenewcommand{\algorithmicensure}{\textbf{Output:}}

\begin{document}
\maketitle
\input{sec/0_abstract}    
\input{sec/1_intro}
\input{sec/2_related_work}
\input{sec/3_method}
\input{sec/4_experiments}
\input{sec/5_conclusion}
\input{sec/6_acknowledgments}
{
    \small
    \bibliographystyle{ieeenat_fullname}
    \bibliography{main}
}

\clearpage
\input{sec/7_appendix}

\end{document}

%% file: sec/0_abstract.tex
\begin{abstract}
 Current language-guided robotic manipulation systems often require low-level action-labeled datasets for imitation learning. While object-centric flow prediction methods mitigate this issue, they remain limited to scenarios involving rigid objects with clear displacement and minimal occlusion.  In this work, we present Embodiment-Centric Flow (EC-Flow), a framework that directly learns manipulation from action-unlabeled videos by predicting embodiment-centric flow. Our key insight is that incorporating the embodiment's inherent kinematics significantly enhances generalization to versatile manipulation scenarios, including deformable object handling, occlusions, and non-object-displacement tasks. To connect the EC-Flow with language instructions and object interactions, we further introduce a goal-alignment module by jointly optimizing movement consistency and goal-image prediction. Moreover, translating EC-Flow to executable robot actions only requires a standard robot URDF (Unified Robot Description Format) file to specify kinematic constraints across joints, which makes it easy to use in practice. We validate EC-Flow on both simulation (Meta-World) and real-world tasks, demonstrating its state-of-the-art performance in occluded object handling (\textbf{62\%} improvement), deformable object manipulation (\textbf{45\%} improvement), and non-object-displacement tasks (\textbf{80\%} improvement) than prior state-of-the-art object-centric flow methods. More results can be found on our project
website: \href{https://ec-flow1.github.io/}{https://ec-flow1.github.io}.
{\let\thefootnote\relax\footnote{$^{\dag}$Corresponding authors.}}
 \end{abstract}

%% file: sec/1_intro.tex
\section{Introduction}
\label{sec:introduction}

Language-guided robotic manipulation has garnered significant attention in recent years. Current state-of-the-art methods have predominantly adopted imitation learning frameworks, where end-to-end Vision-Language-Action (VLA) models have been trained to directly map visual observations and language instructions to low-level robot actions. However, a critical limitation of these approaches stems from their heavy reliance on large-scale datasets annotated with low-level robot action data. Such data are not only difficult to collect at scale but also inherently prone to noise, potentially degrading policy robustness.

\begin{figure}[t]
\centering
\includegraphics[width=\linewidth]{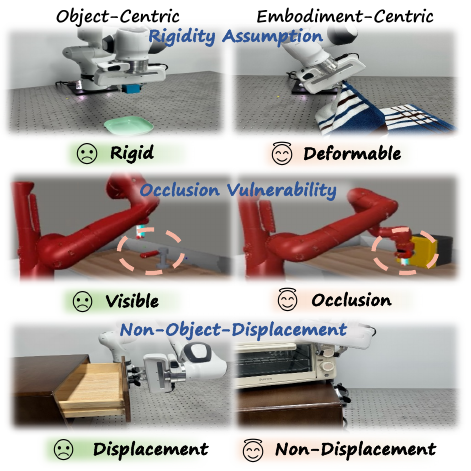}
\caption{Limitations of object-centric flow: (1) Deformable object failure; (2) Struggles with occlusions; (3) Cannot handle non-displacement actions (e.g., rotations or fine motions).}
\label{fig:motivation}
\end{figure}

Beyond limited numbers of action-labeled datasets, abundant manipulation videos without action labels offer rich motion priors that could potentially alleviate data scarcity. Recent works \cite{vip, vlmpc, susie, gr2, dreamitate, gr-mg, subgoal-img-gen} have utilized image and video generation models to infer subgoals for policy learning, improving sample efficiency and generalization. However, these methods still rely on imitation learning, requiring low-level action-labeled data for training. A parallel line of research \cite{tap, im2flow2act, track2act, flip, avdc, general_flow, motion_track} has attempted to bypass action supervision entirely through object-centric flow prediction, where robot motions are inferred from inter-frame transformations of predicted object-centric flow. While promising, these methods have three major limitations in challenging scenarios (Figure \ref{fig:motivation}): (1) \textit{Rigidity assumption}: They presume uniform transformations across object components, failing to handle deformable objects; (2) \textit{Occlusion vulnerability}: Actions are derived solely from object state changes, which makes them ineffective in scenarios with partial observability; (3) \textit{Non-object-displacement manipulation}: They fail to capture actions when objects undergo non-translational state changes (e.g., rotating a round switch) or remain nearly static (e.g., pressing a mouse button) during manipulation. These limitations potentially constrain their generalization to a broader range of real-world manipulation scenarios.

Our approach is motivated by a critical observation: We can fundamentally avoid the limitations by shifting the prediction from object-centric flow to embodiment-centric flow. It eliminates dependence on object properties or states, enabling generalization across deformable and rigid objects, and can also handle occlusions and non-object-displacement manipulation. Moreover, the robot's kinematic structure is visible in most real-world manipulation scenarios – and even under partial occlusion, sufficient motion cues can be extracted from visible joints to infer actions. However, the implementation of embodiment-centric flow poses two significant challenges. (1) Although embodiment-centric flow simplifies the task by emphasizing the robot’s motion dynamics, most manipulation tasks still require precise and meaningful interactions with specific objects. Naive embodiment-centric flow prediction might generate inaccurate robot actions that fail to interact with objects appropriately, especially when language instructions include specific object-related cues. (2) The different joints of the embodiment typically have distinct kinematic constraints such as different ranges of motion, degrees of freedom, and dynamic capabilities, which could not be treated as a unified entity when calculating actions. 

To resolve these issues, we propose a new framework named \textbf{EC-Flow} (\textbf{E}mbodiment-\textbf{C}entric \textbf{F}low) that predicts embodiment-centric flow from action-unlabeled videos with two key designs. First, a goal-conditioned alignment module ensures the predicted flow reconstructs both the robot’s motion trajectory and the goal image specified by language instructions, enabling task-relevant object interactions while adhering to kinematic constraints. Second, a physics-aware action calculation framework leverages the embodiment's URDF (Unified Robot Description Format) file to decompose global EC-flow into joint-specific components via forward kinematics and apply joint-specific transformations, bridging visual predictions with executable actions. We conducted experiments in both the simulation benchmark Meta-World and the real-world scenarios to validate the effectiveness of the method.

Our main contributions can be summarized as follows:
\begin{itemize}
  \item We present the \textbf{EC-Flow} method, which enables manipulation policy learning solely from action-unlabeled video demonstrations. 
  \item We design a goal-image-conditioned flow alignment mechanism that aligns embodiment-centric flow with language instruction and object interactions. 
  \item We propose a URDF-aware action calculating paradigm, which establishes a physics-grounded translation from visual flow predictions to executable actions.
  \item Experimental results demonstrate that EC-Flow effectively handles diverse scenarios, exhibiting strong generalization across deformable objects, occluded environments, and non-object-displacement manipulation tasks. Specifically, it achieves success rate improvements of 62\%, 45\%, and 80\%, respectively, compared to prior state-of-the-art object-centric flow-based methods.
\end{itemize}

%% file: sec/2_related_work.tex
\section{Related Works}
\label{sec:related_work}

\subsection{Imitation Learning}
Imitation learning has been a predominant framework for robotic manipulation, enabling agents to acquire skills through supervised learning from expert demonstrations. While recent Vision-Language-Action (VLA) models \cite{rt1, rt2, octo, openvla, rdt, pi0, spatial_vla, bridgevla} have adopted this paradigm, they rely on large-scale low-level action-labeled datasets. For instance, RDT-1B \cite{rdt} has leveraged over 1 million trajectories (21TB) and OpenVLA \cite{openvla} has utilized the Open X-Embodiment dataset \cite{open_x_embodiment} comprising 2M+ trajectories from 70+ distinct robot platforms. Although such extensive datasets could enable generalizable policies, they have incurred substantial collection costs and introduced challenges including sensor noise aggregation and multi-modal action distributions. These limitations motivate the exploration of action-unlabeled datasets that can circumvent the need for precise low-level action supervision.

\subsection{Learning Manipulation from Videos}
Learning robotic manipulation from action-unlabeled videos leverages abundant, naturally occurring motion priors, eliminating the need for explicit low-level action labels. Traditional methods rely on video prediction \cite{structured, vip, vlmpc, gr1, gr2, dreamitate, unisim} and image generation \cite{gr-mg, subgoal-img-gen} to infer intermediate subgoals, but ultimately depend on action labels for policy learning. 

Another line of work leverages object-centric flow to entirely get rid of the action-labeled dataset, which first predicts object flow and calculates the transformation of the object to infer actions of the object. However, these methods suffer from fundamental scenario-specific limitations: rigidity assumption, occlusion vulnerability, and non-object-displacement manipulation that cannot handle deformable objects, occlusion, or objects with rotational/micro movements scenarios. Among them, General Flow \cite{general_flow} and MT-$\pi$ \cite{motion_track} are most relevant to our approach. In addition to the major difference between object-centric and embodiment-centric, General Flow requires RGBD cameras to predict 3D flow fields, while our EC-Flow uses only RGB videos. MT-$\pi$ relies on tracking 5 predefined gripper keypoints that must remain fully visible during operation, whereas EC-Flow naturally handles partial arm occlusion through embodiment-centric flow prediction. This key difference enables our method to manipulate deformable objects and operate in occluded environments where object-centric approaches fail.

%% file: sec/3_method.tex
\begin{figure*}[t]
\centering
\includegraphics[width=\linewidth]{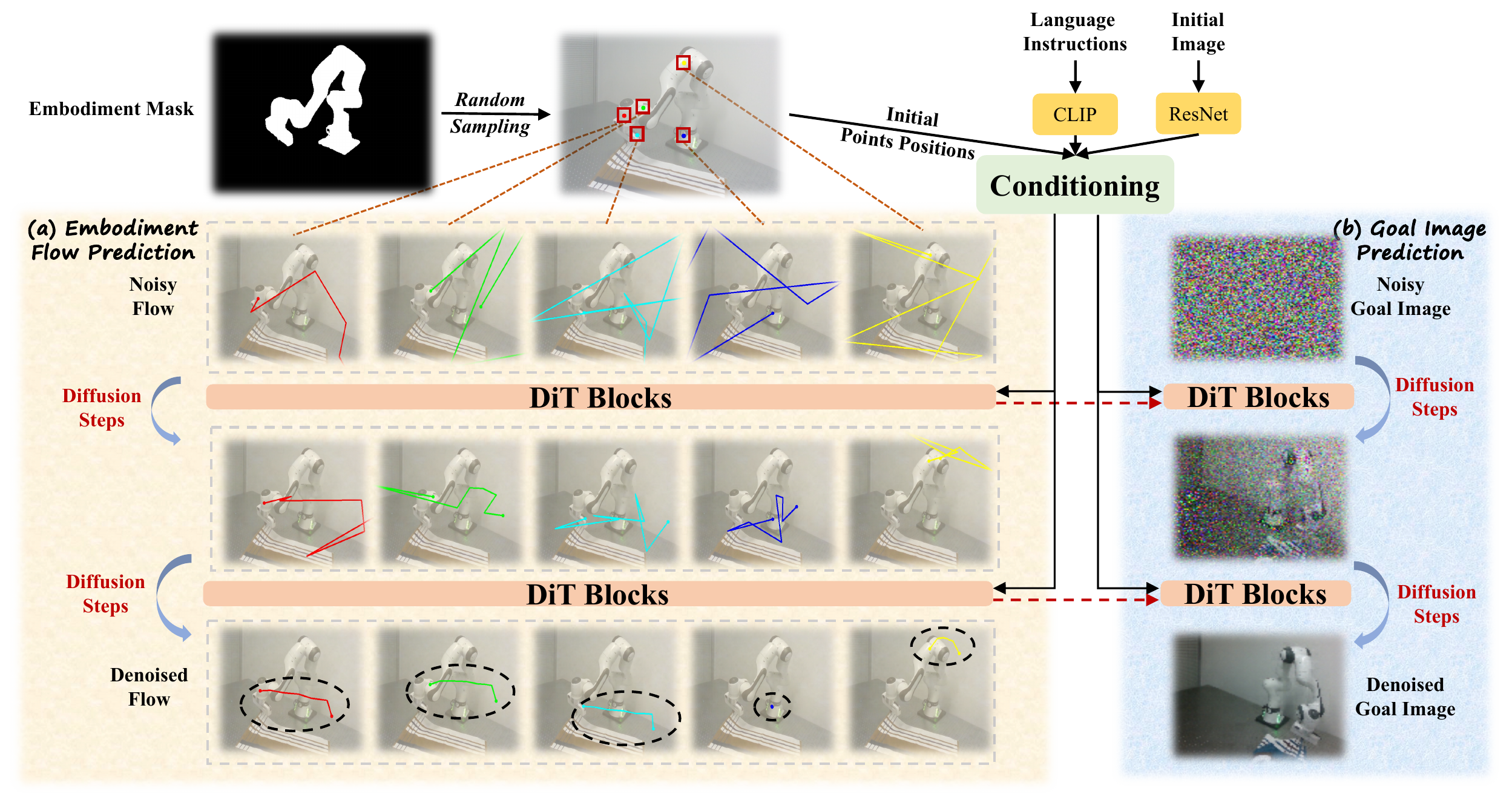}
\caption{Embodiment-centric flow prediction network architecture. \textit{Branch (a)}: prediction of embodiment flow. \textit{Branch (b)}: prediction of the goal image which is used as an auxiliary task for aligning flow to object interactions and language instruction.}
\label{fig:network}
\end{figure*}

\section{EC-Flow Method}
\label{sec:method}
\subsection{Overview}
Our framework addresses language-conditioned manipulation tasks using a video-only dataset $\mathcal{D} = \{\zeta_i\}_{i=1}^n$, where each demonstration $\zeta_i = (\{o_t\}_{t=1}^{m_i}, l_i)$ contains: (1) RGB visual observations $\{o_t\}_{t=1}^{m_i}$ captured from a fixed camera with a total of $m_i$ frames, and (2) a language instruction $l_i$, while explicitly excluding low-level action data.

The proposed \textbf{EC-Flow} overcomes key limitations of prior work through two modules: (1) \textit{Embodiment-Centric Flow Prediction} (Sec \ref{sec: method_flow_pred}): Instead of tracking objects, we predict pixel-wise flow (i.e., future locations) for randomly sampled points of the embodiment; (2) \textit{Kinematic-Aware Action Calculation} (Sec \ref{sec: method_calc_acitons}): Leveraging only the embodiment's URDF model (standard in robotic systems) and consecutive predicted EC-Flow, we compute executable actions without requiring action labels for supervision. This dual-stage approach enables deformable objects, occlusion-robust, and non-object-displacement manipulation purely from RGB video demonstrations.

\subsection{Embodiment-Centric Flow Prediction}

\label{sec: method_flow_pred} To predict embodiment-centric flow, we first construct a flow prediction dataset derived from original RGB videos, then develop a flow prediction model. The primary challenge lies in ensuring that the predicted embodiment-centric flow can effectively capture meaningful interactions between the embodiment and environmental objects for successful instruction execution. To address this fundamental alignment problem, we introduce an auxiliary goal image prediction task, which guarantees that the generated motions adhere to kinematic constraints while also guiding the system toward interactions with objects that are relevant to the instruction. The detailed embodiment-centric flow prediction framework is shown in Figure \ref{fig:network}.

\subsubsection{Dataset Construction}
Let $o_0$ denote the initial video frame. Our pipeline begins with extracting a pixel-wise mask of the embodiment following the structure of Grounded Sam \cite{grounded_sam}, which first obtains a bounding box of the embodiment using text prompt (\textit{"the robotic arm"} in our case) via GroundingDINO \cite{dino} and then refine to pixel-wise masks using SAM2 \cite{sam2} within detected regions. Subsequently, we initialize randomly sampled $N_p$ points within the embodiment mask for tracking. 
We further analyze the sensitivity to the segmentation results in Figure \ref{fig:visualization}, which demonstrates that EC-Flow remains robust to segmentation errors due to the subsequent action calculation procedure.

Ground truth point tracking results are acquired using the off-the-shelf CoTracker \cite{cotracker} model, which simultaneously predicts future \textit{uv} coordinates and visibility of the initial query points across frames. While we save the tracking results across the whole video, our training protocol samples temporal windows of fixed-length $T$ frames from complete trajectories. All coordinates undergo normalization to ensure scale invariance across varying video resolutions. The constructed dataset could be noted as $\mathcal{D} = \{(p_i, o_0^i, l_i)\}_{i=1}^n$, where $p_i \in \mathbb{N}^{N_p \times T \times 3}$.

\subsubsection{Model Architecture}
\textbf{Diffusion-Based Flow Prediction}
\label{sec:dit_flow} Our framework (branch (a) in Figure \ref{fig:network}) implements a conditional diffusion process \cite{diffusion} for flow denoising, formalized through the following Markov chain with discrete timesteps $t \in \{0,\cdots, T\}$,
\begin{equation} q(\mathbf{z}_t|\mathbf{z}_{t-1}) = \mathcal{N}(\mathbf{z}_t; \sqrt{1-\beta_t}\mathbf{z}_{t-1}, \beta_t\mathbf{I}) \end{equation}
where $\mathbf{z}_0$ represents the ground truth flow (\textit{uv} coordinates and visibility states), and $\{ \beta_t\}_{t=1}^T$ defines a cosine noise schedule that gradually corrupts the flow to isotropic Gaussian. The reverse process learns to iteratively denoise the flow through the parameterized transition:
\begin{equation} p_\theta(\mathbf{z}_{t-1}|\mathbf{z}_t) = \mathcal{N}(\mathbf{z}_{t-1}; \mu_\theta(\mathbf{z}_t,t,\mathbf{c}), \Sigma_\theta(\mathbf{z}_t,t,\mathbf{c})) \end{equation} where $\mathbf{c}$ denotes our multimodal conditioning signals. The training objective minimizes the distance between the predicted noise and the ground truth:
\begin{equation}
\mathcal{L}_{\textit{flow}} = \mathbb{E}_{t,\mathbf{z}_0,\epsilon}[\left\|\epsilon - \epsilon_\theta(\mathbf{z}_t,t,\mathbf{c})\right\|_2^2]
\end{equation}
where $\epsilon$ represents the actual noise added at timestep $t$. 

Our multimodal conditioning signals $\mathbf{c}=[\mathbf{\tilde{v}}, \mathbf{\tilde{l}}, \mathbf{\tilde{s}}]$ include three parts: (1) Visual context $\mathbf{\tilde{v}}$: initial frame encoded through ResNet-50 \cite{resnet}; (2) Language guidance $\mathbf{\tilde{l}}$: instruction embeddings from CLIP text encoder \cite{clip}; (3) Starting state $\mathbf{\tilde{s}}$: initial point coordinates. Rather than applying diffusion to the starting point, we treat it as ground truth throughout the generation process by setting the corresponding noise to zero. These heterogeneous features are projected to a unified latent space through learnable adapter layers before cross-attention fusion with trajectory tokens. 

During inference, we employ DDIM \cite{ddim} sampling for efficient trajectory generation. The model output contains the whole trajectory of embodiment point flow during a fixed horizon $T$.

\input{material/algorithm}

\noindent \textbf{Goal Image Prediction}
While the predicted embodiment-centric flow captures motion dynamics, it may not inherently ensure proper object interactions or semantic alignment with language instructions. To bridge this gap, we introduce an auxiliary goal image prediction task that serves as a visual grounding mechanism (branch (b) in Figure \ref{fig:network}).

The goal image generator shares the core architecture of our flow prediction model but operates in the pixel space. During training, we corrupt the goal image (corresponding to the trajectory endpoint) with progressive noise and task the model with reconstructing the denoised image. Crucially, we synchronize the diffusion timesteps between the flow prediction and goal image generation branches, allowing the embodiment flow noise patterns to condition the goal-image generation process. Specifically, at each timestep, the goal-image generator receives augmented conditioning $\mathbf{c_t}^{\textit{img}} = [\mathbf{\tilde{v}}, \mathbf{\tilde{l}}, \mathbf{\tilde{s}}, f_t^{\textit{flow}}]$, where $f_t^{\textit{flow}}$ denotes the flow prediction output at time $t$, creating information transfer between motion estimation and visual synthesis. We denote the loss of the goal image as $L_{\textit{image}}$, and the total loss could be formulated as:
\[
L = L_{\textit{flow}} + \lambda L_{\textit{image}}
\]

This co-training strategy establishes implicit constraints: (1) Object interaction validity: the generated goal image must depict physically plausible object states resulting from the predicted flow; (2) Instruction grounding: visual semantics in the goal image are forced to align with language instructions; (3) Temporal consistency: intermediate flow trajectories must bridge the initial and goal states coherently. By jointly optimizing these complementary objectives, the framework learns to generate the embodiment-centric flow that is both geometrically coherent and semantically aligned with task requirements.

\subsection{Kinematic-Aware Action Calculation}

\label{sec: method_calc_acitons}
Translating predicted flow into executable actions requires addressing the articulated nature of embodiment kinematics. Contrary to prior works that compute unified object transformations, our embodiment-centric flow captures heterogeneous motion patterns across distinct joints, a critical property demanding specialized calculation mechanics.

Our kinematic-aware calculation pipeline (Algorithm~\ref{alg:action_gen}) operates through two core phases, we first allocate sampled points to joints and then calculate the transformation of the end-effector.

\subsubsection{Allocate Sampled Points to Joints}
We begin by filtering the sampled points according to three criteria: (1) consistent visibility across consecutive frames, (2) significant displacement surpassing a minimum motion threshold, and (3) valid depth information from the camera.

\begin{figure}[h]
\centering
\includegraphics[width=\linewidth]{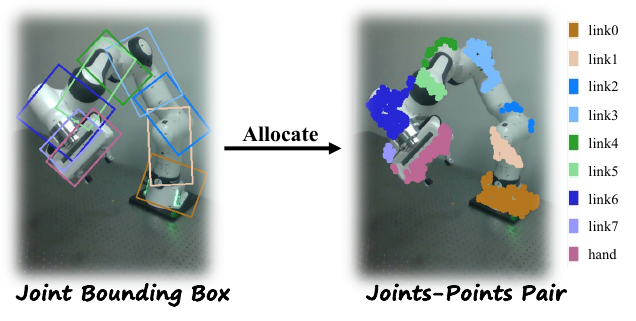}
\caption{Process of allocating sampled points to specific joints. We only select points that only belong to a single joint.}
\label{fig:assign_points}
\end{figure}

For joint assignment (shown in Figure \ref{fig:assign_points}), we first utilize the geometric attributes provided in the URDF file to determine the 3D positions of the joints. Subsequently, the corresponding 2D bounding boxes of joints on the image plane are computed using the intrinsic and extrinsic parameters of the camera. A point is considered to belong to a joint if it lies within the computed bounding box. Due to the potential overlap of joints in the 2D observation, only points uniquely assigned to a single joint are selected for further processing.

\subsubsection{Calculate Transformation of End-Effector}
Given the established joint-point correspondences, we compute executable end-effector actions. The pipeline involves three core operations:

(1) Depth-aware Reconstruction: Lift the 2D point flow to 3D coordinates using calibrated depth maps and camera parameters. (2) Joint Transformation Calculation: For candidate end-effector poses, compute associated joint transformations via URDF-based inverse kinematics. (3) Reprojection Alignment: Optimize the proposed end-effector pose by minimizing the discrepancy between (a) predicted embodiment flow trajectories and (b) reprojected trajectories generated from the estimated poses, ensuring consistency through iterative refinement.

Specifically, given a target end-effector pose $T_{ee}$, we first compute each joint's transformation relative to end-effector $_{j}^{ee}\mathbf{T}$ via inverse kinematics, and compute the reprojection error:
\[
\epsilon = \sum\limits_{j=1}^M\sum\limits_{i=1}^{N_j}{\left\| \pi(\mathbf{T}_{ee} \cdot _{j}^{ee}\mathbf{T} \cdot \mathbf{P}_{ji}^{(t)_{3D}})-\mathbf{P}_{ji}^{{(t+1)}_{2D}} \right\|_2}
\]
where $\pi(\cdot)$ denotes camera projection. 

The optimal pose $\mathbf{T}_{ee}^*$ is obtained through the minimization of $\epsilon$, constrained by the embodiment's mechanical limits. This geometrically grounded approach ensures physical plausibility while maintaining alignment with visual observations.  The resulting optimal end-effector pose, $\mathbf{T}_{ee}^*$, is then transmitted to the control systems for action execution.

%% file: material/algorithm.tex
\begin{algorithm*}[t]
\caption{Kinematic-Aware Action Calculation}
\label{alg:action_gen}
\begin{algorithmic}[1]
\Require 
\Statex $\{\mathcal{P}^{(t)_{2D}}\}_{t=0}^{T}$: predicted 2D EC-Flow of horizon $T$,  
$\mathcal{K}$: camera intrinsic/extrinsic parameters, 
$\mathcal{J}$: embodiment joint configuration, 
$\pi(\cdot)$: camera projection function, 
$M$: number of joints, $\{\mathbf{T}_{ee}^{(t)}\}_{t=0}^{T-1}$: pose matrix of the end-effector to be optimized.
\Ensure 
\Statex $\{\mathbf{T}_{ee}^{*(t)}\}_{t=0}^{T-1}$: optimal pose matrix of the end-effector for each time step
\For{$t = 0$ to $T - 1$}
    \State $\mathcal{P}_{filt}^{(t)_{2D}} \gets \text{FilterPoints}(\mathcal{P}^{(t)_{2D}})$ 
    \Comment{Filter points for stable calculation}
    
    \State $\{\mathbf{b}_j\}_{j=1}^M \gets \text{ProjectJointPositions}(\mathcal{J}, \mathcal{K})$ 
    \Comment{Get 2D joint bounding box using URDF geometry}
    
    \State $\{\mathcal{P}_j^{(t)_{2D}}\}_{j=1}^M \gets \text{AssignPointsToJoints}(\mathcal{P}_{filt}^{(t)_{2D}}, \{\mathbf{b}_j\}_{j=1}^M)$  
    \Comment{Establish joint-point correspondences}

    \State $\{\mathcal{P}_j^{(t)_{3D}}\}_{j=1}^M \gets \text{LiftTo3D}(\{\mathcal{P}_j^{(t)_{2D}}\}_{j=1}^M, \textit{Depth}, \mathcal{K})$ 
    \Comment{Depth-aware reconstruction}

    \State $\{_{j}^{ee}\mathbf{T}^{(t)}\}_{j=1}^M \gets \text{ComputeJointTransforms}(\mathbf{T}_{ee}^{(t)}, \mathcal{J})$ 
    \Comment{Compute joint transforms for a given $\mathbf{T}_{ee}^{(t)}$ via IK}

    \State $\mathbf{T}_{ee}^{*(t)} \gets \arg\min_{\mathbf{T}_{ee}^{(t)}} \sum\limits_{j=1}^M\sum\limits_{i=1}^{N_j}{\left\| \pi(\mathbf{T}_{ee}^{(t)} \cdot _{j}^{ee}\mathbf{T}^{(t)} \cdot \mathbf{P}_{ji}^{(t)_{3D}})-\mathbf{P}_{{ji}}^{{(t+1)}_{2D}} \right\|_2}$ 
    \Comment{Optimization of the EEF pose at time $t$}
\EndFor
\end{algorithmic}
\end{algorithm*}

%% file: sec/4_experiments.tex
\input{material/table/simulation}
\section{Experiment Results}
\label{sec:experiment}
\subsection{Simulation Evaluation}
\textbf{Simulation Setup}~~~~
We conduct our experiment within the Meta-World simulation benchmark \cite{metaworld}, which provides a simulated environment for a series of tabletop manipulation tasks using a Sawyer robot arm. This benchmark involves interactions with various objects and tools, where the object locations are randomly initialized at the start of each task. In the Meta-World setup, the robot's gripper is assumed to remain in a downward orientation throughout, resulting in an action space with four dimensions: the position of the end effector and the gripper's open/close state.

We evaluated nine tasks from the Meta-World benchmark. For each task, we used a side-mounted camera in the environment to capture five video demonstrations, resulting in a total of 45 videos. The policy is evaluated 25 times for each task with randomly initialized object positions. Success is defined as reaching a pre-defined goal within a predefined maximum number of steps. We trained our EC-Flow policy using 8 Nvidia 4090 GPUs for one day.

\noindent \textbf{Compared Methods}~~~~
We compare our method against six state-of-the-art methods:
\begin{enumerate}
\item [(1)] BC-Scratch: Standard behavior cloning baseline with dual-stream encoder architecture, which processes RGB observations through ResNet-18 \cite{resnet} and language instructions via CLIP \cite{clip} and is trained end-to-end on expert demonstrations.
\item [(2)] BC-R3M \cite{r3m}: Enhances visual representation learning by initializing the image encoder with R3M pretrained weights, while maintaining identical architecture and training protocol as BC-Scratch.
\item [(3)] UniPi \cite{unipi}: Two-stage approach combining: (i) vision-language conditioned video prediction model, and (ii) inverse dynamics model that maps predicted frames to actions through behavioral cloning.
\item [(4)] Diffusion Policy \cite{diffusion_policy}: Diffusion-based action predictor that models action sequences as denoising processes conditioned on observation-history windows, trained via imitation learning.
\item [(5)] AVDC \cite{avdc}: Object-centric flow paradigm that first learns video prediction, then extracts optical flow trajectories using off-the-shelf point trackers, and finally computes action through object flows.
\item [(6)] Track2Act \cite{track2act}: Approach that directly uses object-centric flow prediction. Unlike the original paper (incorporating an extra module using action-labeled data to refine actions), we here implement the approach for computing actions solely from the flow.
\end{enumerate}

\noindent \textbf{Implementation Details}~~~~
Our grasping strategy adopts the object-centric approach from \cite{avdc}, 
the grasp point is the centroid of the segmented mask.
For motion planning robustness, we also implement an adaptive replanning protocol, as described in \cite{avdc}, which is triggered when the actions do not change beyond a given threshold. 

For EC-Flow, we predict $T=8$ steps over $N_p=400$ sampled points. Both the start and goal images are resized to $128 \times 128$.
We perform $250$ sampling iterations for DDIM during the diffusion inference process. The batch size is set to $56$, with a learning rate of $5\times10^{-5}$ for the flow prediction and $1\times10^{-4}$ for the goal image prediction. To balance the contributions of the flow prediction and the goal image prediction, we set $\lambda$ to $0.4$.

\noindent \textbf{Results}~~~~
The simulation results on the Meta-World benchmark are presented in Table \ref{tab:metaworld}. Methods (1)–(4) rely on traditional imitation learning using low-level action-labeled datasets, while methods (5)–(6) utilize only action-unlabeled video data. Specifically, method (5) employs video prediction followed by flow tracking, whereas method (6) directly predicts object-centric flow.

\begin{figure}[h]
\centering
\includegraphics[width=\linewidth]{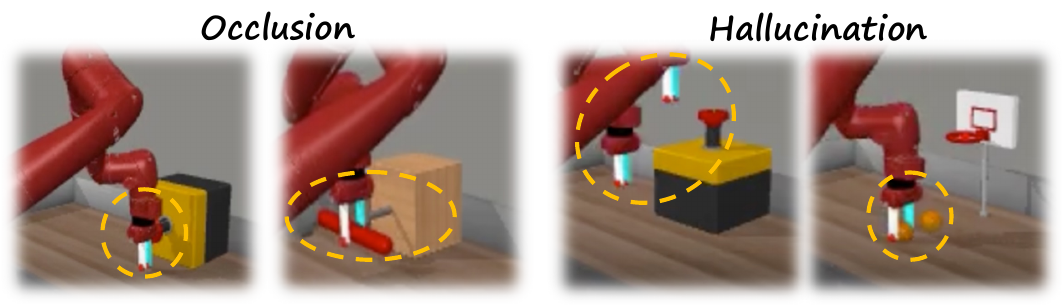}
\caption{Occlusion and hallucination problems of object-centric flow and video prediction methods.}
\label{fig:prior_drawback}
\end{figure}

The results demonstrate a significant improvement, particularly in the \textit{btn-top-press} and \textit{hammer-strike} tasks, where our approach achieves an average success rate increase of \textbf{65\%} compared to prior object-centric flow methods. Moreover, our method surpasses previous state-of-the-art approaches on the overall Meta-World benchmark by \textbf{16.4\%}. This improvement is primarily attributed to the challenges posed by object occlusions in these tasks, where previous methods struggle with accurately tracking occluded objects, often leading to failure. As illustrated in Figure \ref{fig:prior_drawback}, prior methods fail because of occlusion of target objects or hallucination effects introduced by video prediction, which can lead to erroneous object tracking. In contrast, our EC-Flow method effectively handles such cases, even when parts of the embodiment remain unobservable.

\noindent \textbf{Ablation Study}~~~~
We systematically evaluate four critical design variants to isolate component contributions. The \textit{Video+GT Flow} configuration replaces our end-to-end flow prediction with flow estimates from a pre-trained video prediction model. Removing the goal image prediction auxiliary task (\textit{Goal Img Pred}) tests multimodal alignment necessity. For action generation, disabling the point filtering strategy (\textit{Point Filter}) before action calculation reveals motion reliability criteria importance while restricting tracked points to the end-effector (\textit{EEF Points}) evaluates full-arm motion modeling benefits. 

Model \#2 shows that combining a video prediction model with an off-the-shelf point tracking method performs worse than our end-to-end EC-Flow approach. As illustrated on the right side of Figure \ref{fig:prior_drawback}, video prediction can suffer from hallucination issues—predicting multiple robotic arms—which then introduces errors in point tracking. This problem is avoided by using end-to-end flow prediction. Model \#3 highlights the necessity of the goal-image prediction module; directly predicting embodiment-centric flow may be misaligned with object interactions or language instructions, whereas the additional module alleviates this issue. Model \#4 demonstrates the effectiveness of the point filtering strategy since static, invisible, or invalid-depth points can degrade the overall performance of the action calculation process. Finally, model \#5 underscores the importance of computing the complete set of embodiment points. In scenarios where the end-effector is occluded, considering the full embodiment helps reduce errors.

\input{material/table/ablation}

We also evaluate various configurations of the sample point count $N_p$ and the prediction horizon $T$ in the EC-Flow prediction network. Our results indicate that using fewer sample points degrades performance, as inaccuracies from outlier points impair the action calculation. Additionally, a shorter prediction horizon may miss critical stages of the manipulation process, and larger gaps between consecutive actions can compound errors during calculation, further reducing performance.  Increasing $N_p$ and $T$ does not lead to substantial differences, we select the current configuration to balance performance with training and inference costs.

\begin{figure}[ht]
\centering
\begin{minipage}[c]{0.48\linewidth}
    \centering
    \includegraphics[width=\textwidth]{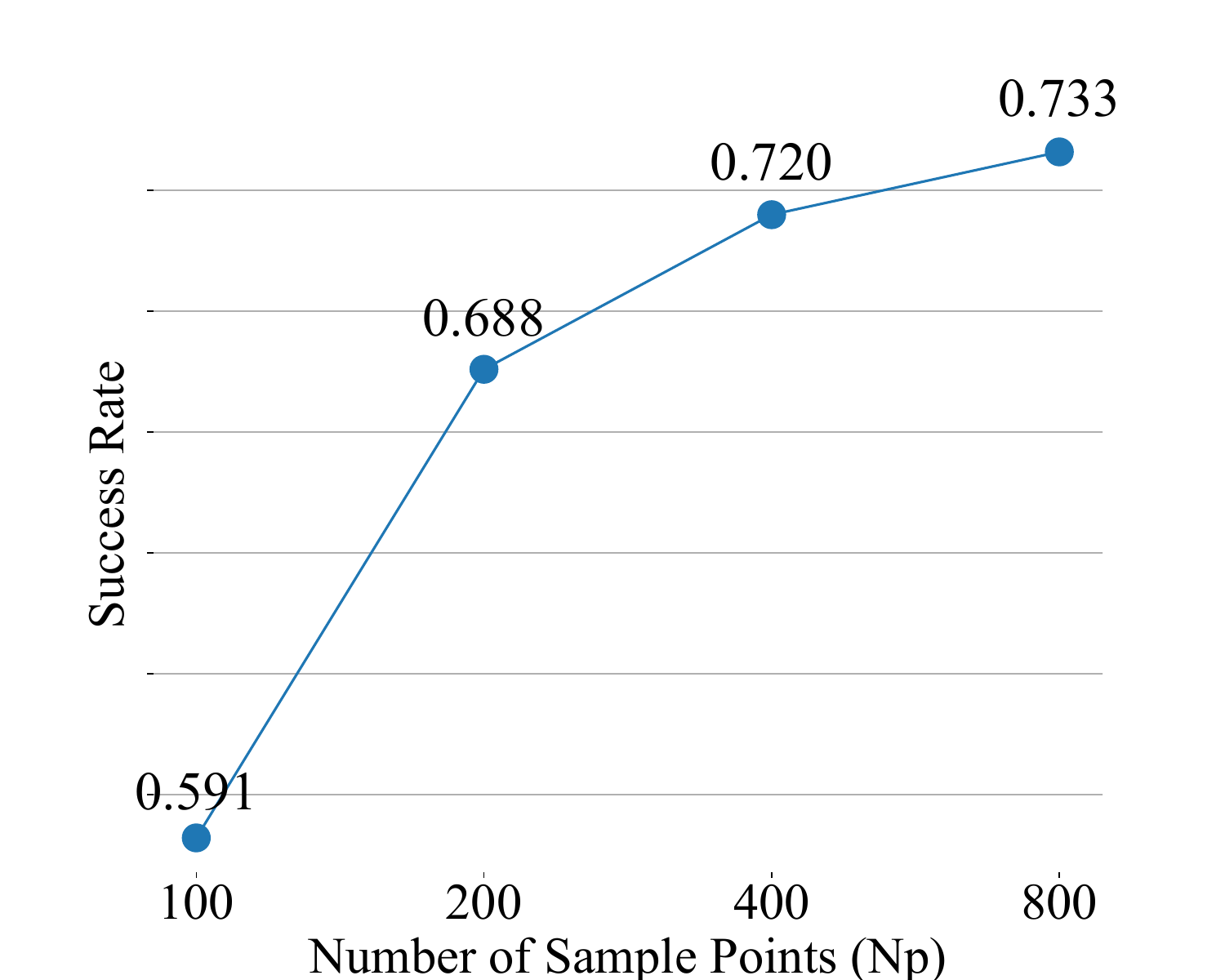}
\end{minipage}
\begin{minipage}[c]{0.48\linewidth}
    \centering
    \includegraphics[width=\textwidth]{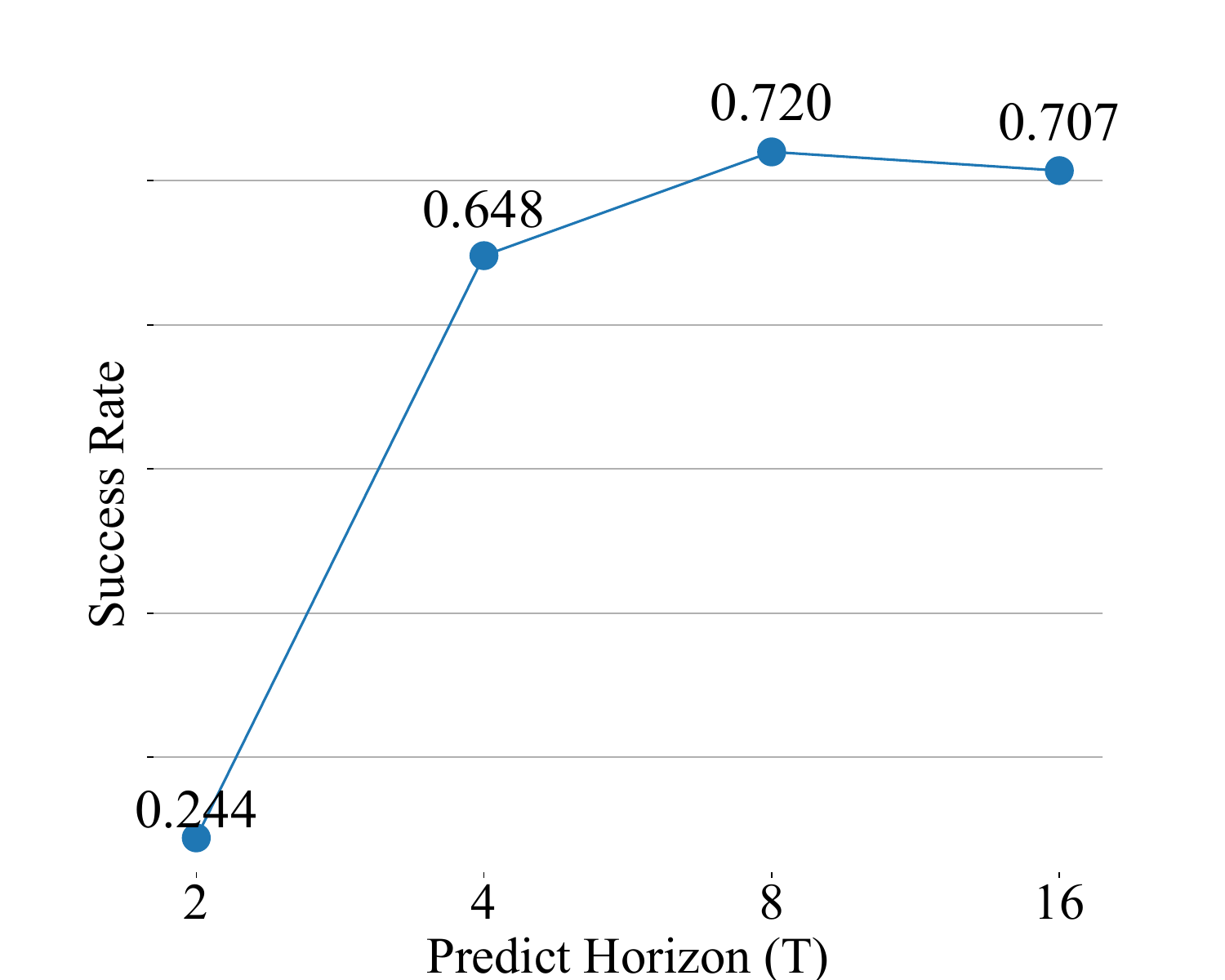}
\end{minipage} 
\caption{\textit{Left}: Influence of choice of sampled points $N_p$.  \textit{Right}: Influence of choice of predict horizon $T$.}
\label{fig:time_camparison}
\end{figure}

\input{material/table/realworld}

\begin{figure*}[t]
\centering
\includegraphics[width=\linewidth]{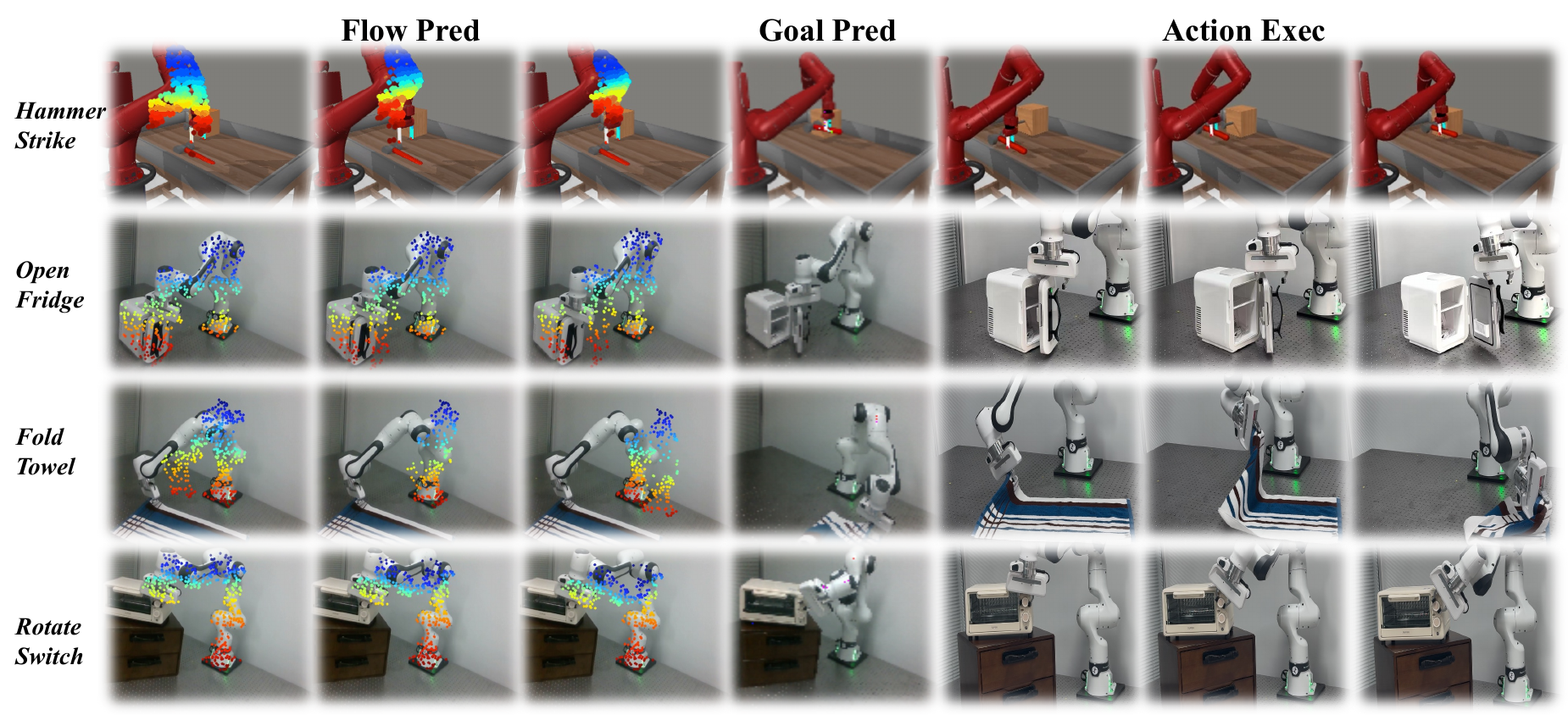}
\caption{Visualization of the EC-Flow method for learning robotic manipulation from action-unlabeled video data. \textit{Left} (3 cols): flow prediction results. \textit{Middle} (1 col): goal image prediction results. \textit{Right} (3 cols): final action execution results. View differences originate from distinct camera positions for data capture (\textit{Left/Middle}) and execution recording (\textit{Right}).}
\label{fig:visualization}
\end{figure*}

\subsection{Real-world Evaluation}
\textbf{Real-world Setup}~~~~
Our real-world experiment setup consists of a statically mounted Franka Research 3 robotic arm and a static third-person view Intel Realsense D435i RGB-D camera.  We calibrate the robot-camera extrinsic and convert the perceived 3D points into the robot's base frame before calculating actions. For a given target gripper action, we utilize Deoxys \cite{deoxys} to guide the robot to the target through trajectory generation and feedback control.

Seven manipulation tasks are constructed across three categories: (a) rigid objects with prismatic/revolute joints (3 tasks), (b) deformable objects (2 tasks), and non-object-displacement manipulation (2 tasks). Each task is trained on five action-unlabeled video demonstrations.

We adopt two baselines: Behavior Cloning (BC) using action-labeled data and Track2Act\cite{track2act} using object-centric flow, which is consistent with the simulation baselines. For all baselines, we manually position the robotic arm for task
initiation and we test each task 10 times with objects of randomly initialized positions.

\noindent \textbf{Results}~~~~The results of the real-world evaluation are presented in Table \ref{tab:realworld}. The results of the real-world evaluation are presented in Table \ref{tab:realworld}. Our findings show that EC-Flow achieves a \textbf{45\%} improvement in deformable object manipulation and an \textbf{80\% }improvement in non-object-displacement manipulation tasks compared to prior object-centric flow-based methods, which are incapable of handling these tasks. 

Moreover, our method surpasses behavior cloning approaches that depend on action-labeled data for supervision. These methods typically require a large number of expert demonstrations to learn an effective policy, making them less effective in data-limited scenarios. In contrast, our approach achieves superior performance with only five videos per task. This is enabled by leveraging embodiment-centric flow in the observation space, allowing EC-Flow to generalize more effectively to new objects and robot positions without relying on low-level action supervision.

The relatively low success rate in the \textit{fold clothes} task may be partially attributed to inaccuracies of depth perception from the low-cost D435i camera. The action accuracy could potentially be improved by using a sensor with more precise depth-sensing capabilities.

\noindent \textbf{Qualitative Analysis}~~~~
We visualize our EC-Flow method in Figure \ref{fig:visualization}. The process begins with utilizing goal image prediction to assist in generating the embodiment-centric flow, followed by directly computing actions from the flow. In the second row of Figure \ref{fig:visualization}, we further demonstrate the robustness of our method to inaccuracies in segmentation results produced by off-the-shelf models. Specifically, in that case, the segmentation model mistakenly identifies a part of the fridge as part of the embodiment. However, our method effectively filters out these erroneous points by leveraging the spatial positions of the robotic arm’s joints, as defined in the URDF file. We demonstrate that EC-Flow is capable of handling deformable objects and performing non-object-displacement manipulation.

%% file: material/table/simulation.tex
\begin{table*}[t] 
    \tiny
    \centering
    \begin{adjustbox}{width=\textwidth}
    \begin{tabular}{@{}lcccccccccc@{}}
    \toprule
    & Avg. & door & door & shelf & btn & btn-top & faucet & faucet & handle & hammer\\
    & Success $\uparrow$ & open & close & place & press & press & close & open & press & strike \\
    \midrule
    BC-Scratch~ & 0.204 & 0.24 & 0.36 & 0 & 0.36 & 0.12 & 0.20 & 0.20 & 0.36 & 0\\
    BC-R3M~\cite{r3m} & 0.360 & 0.04 & 0.60 & 0 & 0.36 & 0.04 & 0.24 & 0.68 & 0.52 & 0.76\\
    UniPi~\cite{unipi} & 0.093 & 0 & 0.36 & 0 & 0.12 & 0 & 0.04 & 0.12 & 0.16 & 0.04\\
    Diffusion Policy~\cite{diffusion_policy} & 0.298 & 0.48 & 0.48 & 0 & 0.40 & 0.20 & 0.24 & 0.60 & 0.24 & 0.04\\
    AVDC~\cite{avdc} & 0.489 & 0.72 & 0.92 & \textbf{0.20} & 0.60 & 0.24 & \textbf{0.56} & 0.24  & \textbf{0.84} & 0.08 \\
    Track2Act~\cite{track2act} & 0.556  & 0.88 & 0.76 & 0.04 & 0.56 & 0.40 & 0.48 & 0.88 & 0.76 & 0.24\\
    EC-Flow(ours)~ & \textbf{0.720} & \textbf{0.96} & \textbf{1.00} & 0.04 & \textbf{0.64}  & \textbf{1.00} & 0.44 & \textbf{1.00}  & 0.52  & \textbf{0.88}  \\
    \bottomrule
    \end{tabular}
    \end{adjustbox}
    \caption{Simulation results on Meta-World benchmark.}
\label{tab:metaworld}
\end{table*}

%% file: material/table/ablation.tex
\begin{table}[h]

    \small
    \centering
    \begin{adjustbox}{width=0.7\linewidth}
    \begin{tabular}{c|cc|cc|c}
     \toprule
    \multirow{3}{*}{\textbf{\#}} & \multicolumn{2}{c}{Flow Pred}\vline& \multicolumn{2}{c}{Action Calc}\vline& \multirow{3}{*}{\textbf{Avg} }\\ \cmidrule{2-3} \cmidrule{4-5} 
                                        & {Video} & {Goal Img} & {Point}
                                        & {EEF}\\ 
                                        & {+ GT Flow} & {Pred} & {Filter}
                                        & {Points}\\ 
                                        \midrule \midrule
1 & - &$\checkmark$   & $\checkmark$   & -  & \textbf{0.720}\\
2 & - & - &  $\checkmark$    &  - & 0.636 \\ 
3 & - &$\checkmark$    & -  & -  & 0.582   \\ 
4 & - &$\checkmark$   & $\checkmark$   & $\checkmark$ & 0.604   \\
5 & $\checkmark$   & $\checkmark$    & $\checkmark$ & -  & 0.667   \\ 
\hline
    \end{tabular}
    \end{adjustbox}
  \caption{Ablation analysis of EC-Flow design.}
  \label{tab:abla}
\end{table}

%% file: material/table/realworld.tex
\begin{table*}[t] 
    \tiny
    \centering
    \begin{adjustbox}{width=0.8\textwidth}
    \begin{tabular}{@{}lcccccccccc@{}}
    \toprule
    & All & open & open & open & fold & fold & press & rotate\\
    & Tasks $\uparrow$ & fridge  & drawer & oven & clothes & towel & mouse & switch \\
    \midrule
    BC~ & 31/70 & 5/10 & 6/10  & 6/10 & \textbf{4/10} & 2/10  & 3/10  & 5/10\\
    Track2Act~\cite{track2act} & 21/70 & 8/10 & 8/10  & 5/10 & 0/10 & 0/10  & 0/10  & 0/10\\
    EC-Flow(ours)~ & \textbf{54/70} & \textbf{9/10} & \textbf{10/10}  & \textbf{10/10} & 3/10 & \textbf{6/10}  & \textbf{9/10}  & \textbf{7/10}  \\
    \bottomrule
    \end{tabular}
    \end{adjustbox}
    \caption{Results on real-world manipulation tasks.}
\label{tab:realworld}
\end{table*}

%% file: sec/5_conclusion.tex
\section{Conclusion}
\label{sec:conclusion}
We propose EC-Flow, a framework that learns embodiment-centric flow directly from action-unlabeled videos. By jointly predicting flow and goal images, it captures both the robot’s motion and object interactions. Moreover, actions are computed using only a standard URDF file, making the method easy to deploy. Experiments in simulation and the real world show that EC-Flow effectively handles diverse manipulation tasks, including deformable objects, occlusions, and non-object-displacement scenarios.

%% file: sec/6_acknowledgments.tex
\section{Acknowledgments}
This work was jointly supported by National Natural Science Foundation of China (62322607, 62236010 and 62276261), and Youth Innovation Promotion Association of Chinese Academy of Sciences under Grant 2021128.

%% file: sec/7_appendix.tex
\maketitlesupplementary
\appendix

\section{Limitations and Future Work}
EC-Flow performs manipulation starting from a manually set initial pose and computes the subsequent actions. Future work may involve integrating the off-the-shelf grasping pose generation models \cite{graspnet, dexgrasp} to facilitate starting position initialization, thereby enhancing overall task efficiency. 

Additionally, we aim to leverage foundation multi-modal models to extract gripper state information directly from the video dataset, which could then be integrated into the flow prediction network to facilitate gripper state prediction.

\section{Design Choices of EC-Flow}

\subsection{Trade-offs between EC-Flow and End-Effector-Regression-Only Methods}
End-effector-regression methods refer to those that directly regress the end-effector (EEF) pose using re-projection error, without optimizing the full joint configuration. Table~\ref{tab:inference_speed} compares the inference latency and task success rates of different methods. For our EC-Flow approach, flow prediction requires 4.37\,s for 8 frames—performed only once at the start of each trajectory and not involved during execution. This step could be further accelerated by replacing diffusion with flow matching. The point projection step takes 0.01\,s, while action computation takes 0.21\,s when regressing only the EEF pose, or 0.37\,s when performing full-joint optimization, due to the added cost of inverse kinematics and optimization complexity.

In terms of task performance, the EEF-only baseline shows a drop in success rates by 5.3\% in simulation and 7.1\% in the real world, while achieving 1.76$\times$ faster inference. The performance drop in simulation is mainly due to the \textit{door-open} task, where the EEF becomes heavily occluded. In the real world, it stems from the \textit{fold-towel} task, where the EEF shifts direction significantly—from the side to the front—making it difficult to track the initial side-view points. To address these challenges, we adopt full-joint optimization, which helps mitigate occlusions and better accommodate large directional changes of the EEF, at a moderate computational cost. Additionally, joint-specific weights in the re-projection error can be manually tuned based on joint visibility or task requirements. To improve optimization convergence and stability, we initialize from the previous pose estimate.

\input{material/table/inference_speed}

\subsection{Goal Image VS. Object Flow Prediction}
We initially explored object flow to support EC-Flow by modeling object interaction, but found it struggles to converge on deformable objects due to unstructured motion patterns (such as folding towel), while goal image converges faster and could also serve as a proxy for modeling interaction. However, we agree that for complex tasks with multiple required intermediate object states, object flow could be more useful than using goal-image alone. 

\subsection{Sensitivity to Camera Pose Selection}
Our EC-Flow method does not rely on a specific camera pose that captures all robot joints in view. In practice, having visibility of just 2–3 joints is sufficient for reliable action computation. EC-Flow allows flexible inclusion of any visible joints in the optimization, adapting dynamically based on their visibility.

However, it is crucial that the set of visible joints remains consistent across frames. Since EC-flow is predicted only from the initial frame, if a joint visible at the beginning becomes fully occluded later, the system may fail to track it, potentially leading to execution failure.

\section{Use of Internet Cross-Embodiment Data}
To investigate the potential of cross-embodiment video data under limited robot demonstrations, we conduct a preliminary study on two representative tasks—\textit{door-open} and \textit{door-close}—from the Meta-World benchmark~\cite{metaworld}. We augment a small set of robot demonstrations with 50 human videos of the same tasks sourced from the Something-Something-v2 dataset~\cite{sth-sth-v2}.

We evaluate in a low-data regime with only 2 robot demonstrations per task (compared to 5 in the standard setting). As shown in Table~\ref{tab:cross_embodiment}, human video data alone fails to achieve zero-shot transfer to the robot embodiment. Using 2 robot demos achieves a 46\% success rate; adding 5 human videos improves this to 52\%, while incorporating 50 human videos further boosts performance to 70\%.

These results suggest that human videos provide valuable motion priors and can significantly enhance sample efficiency in robot learning. We believe that large-scale pretraining on internet-scale human video datasets, followed by robot-specific fine-tuning, represents a promising direction for future research.

\input{material/table/cross_embodiment}

\section{Implementation Details}
\subsection{Real-World Setup}
\begin{figure}[h]
\centering
\includegraphics[width=0.8\linewidth]{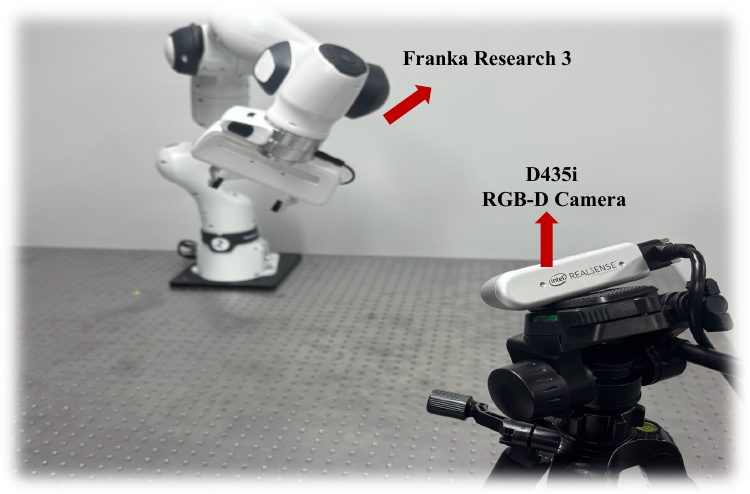}
\caption{The real-world setup of VERM.}
\label{fig:realworld_setup}
\end{figure}

Our real-world setup is illustrated in Figure \ref{fig:realworld_setup}. 

\subsection{Real-world Baselines}
\begin{enumerate}
\item [(1)] BC: Standard behavior cloning baseline with dual-stream encoder architecture. Processes RGB observations through ResNet-18 \cite{resnet} and language instructions via CLIP \cite{clip}, trained end-to-end on action-labeled demonstrations.

\item [(2)] Track-2-Act \cite{track2act}: An object-centric flow prediction method that predicts object flow and calculates actions from object movements.
\end{enumerate}

\subsection{Parameter Details}
The parameters of the flow prediction and goal-image prediction network are shown in Table \ref{tab:flow_pred_param} and Table \ref{tab:goal_image_pred_param} respectively.

\begin{table}[ht]
\centering
\small
\begin{tabular}{cc}
\midrule
Parameter & Value \\ 
\midrule
horizon & 8 \\ 
num\_points & 400 \\ 
img\_size & 128 \\ 
num\_sampling\_steps & 250 \\ 
batch\_size & 56 \\ 
optimizer & AdamW \\ 
lr & 5.0e-5 \\ 
loss\_weight & 1 \\
transformer depth & 24 \\
transformer hidden\_size & 1152 \\
transformer num\_heads & 16 \\
mlp\_ratio & 4.0 \\
lang\_dim & 1024 \\
img\_dim & 512 \\
\bottomrule
\end{tabular}
\caption{Hyper-parameters of flow prediction network.}
\label{tab:flow_pred_param}
\end{table}


\begin{table}[ht]
\centering
\small
\begin{tabular}{cc}
\midrule
Parameter & Value \\ 
\midrule
horizon & 8 \\ 
num\_points & 400 \\ 
img\_size & 128 \\ 
num\_sampling\_steps & 250 \\ 
batch\_size & 56 \\ 
optimizer & AdamW \\ 
lr & 1.0e-4 \\ 
loss\_weight & 0.4 \\
transformer depth & 12 \\
transformer hidden\_size & 384 \\
transformer num\_heads & 6 \\
patch\_size & 16 \\
lang\_dim & 1024 \\
img\_dim & 512 \\
\bottomrule
\end{tabular}
\caption{Hyper-parameters of goal image prediction network.}
\label{tab:goal_image_pred_param}
\end{table}

\section{More Visualization Results}
\begin{figure*}[t]
\centering
\includegraphics[width=\linewidth]{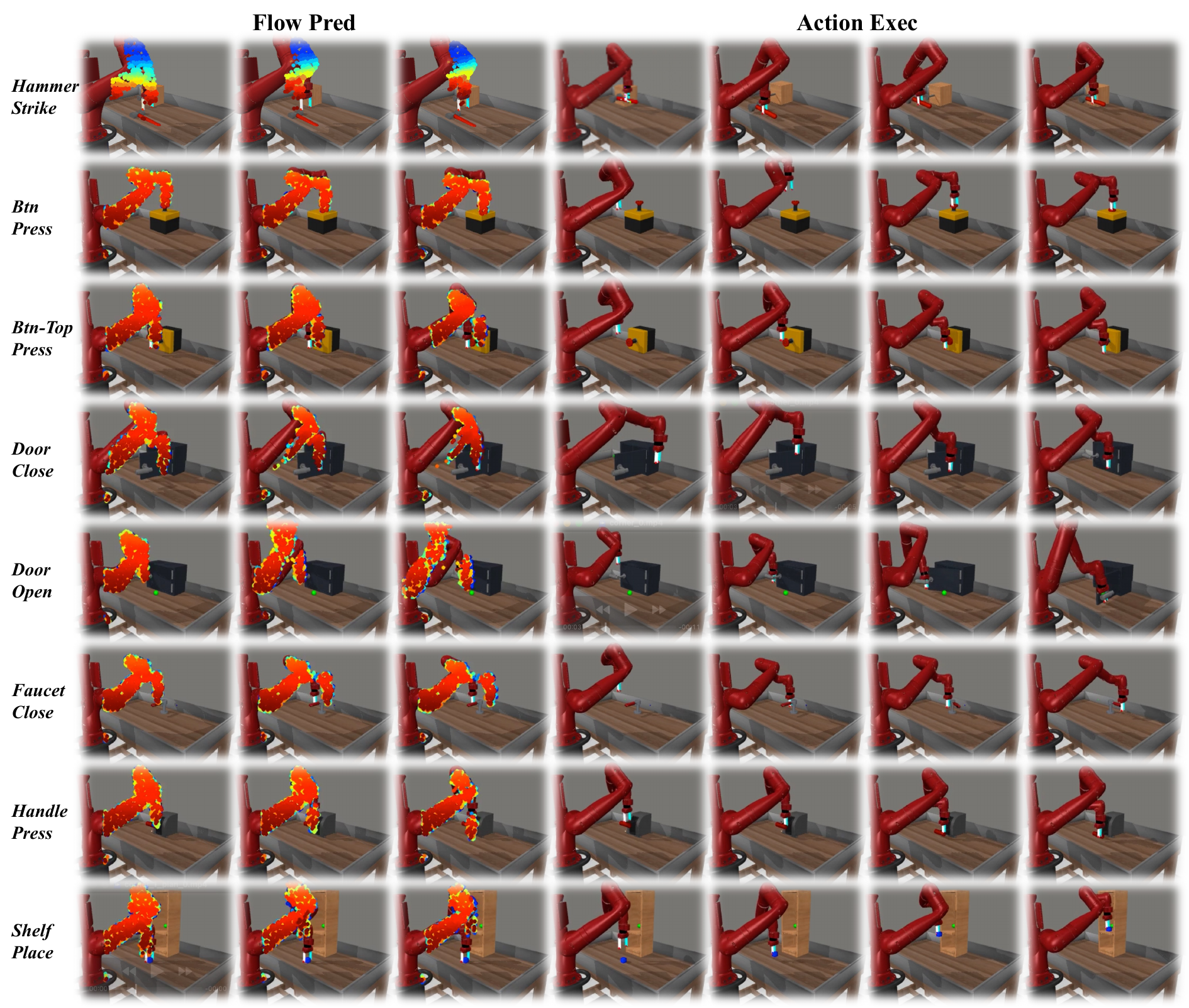}
\caption{Visualization of the EC-Flow in Meta-World.}
\label{fig:vis_meta_all}
\end{figure*}

We demonstrate more visualization results of EC-Flow in the Meta-World (Figure \ref{fig:vis_meta_all}) and real-world (Figure \ref{fig:vis_real_all}). The demonstration videos can be found in the supplementary video materials.

\begin{figure*}[t]
\centering
\includegraphics[width=\linewidth]{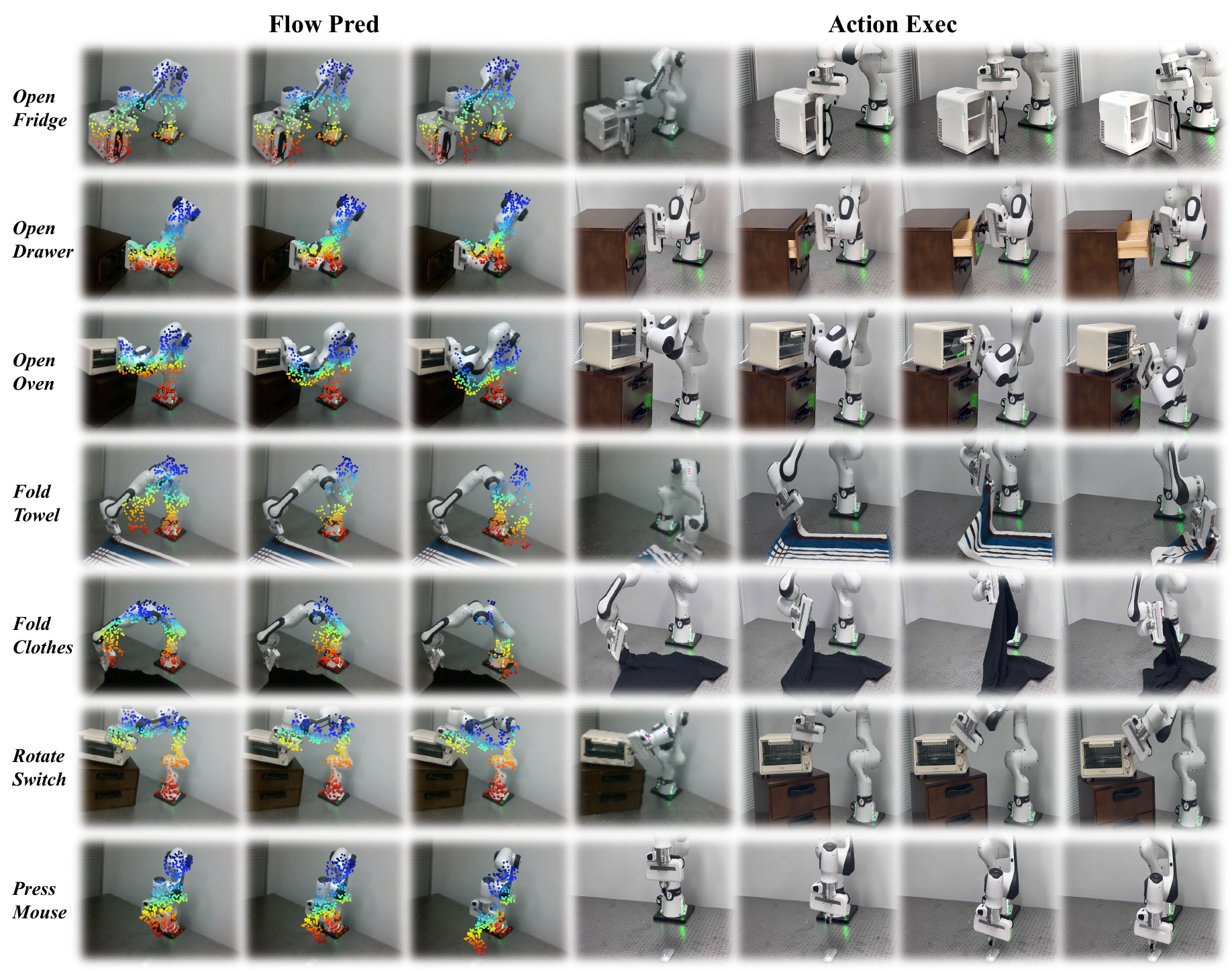}
\caption{Visualization of the EC-Flow in the real-world.}
\label{fig:vis_real_all}
\end{figure*}

%% file: material/table/inference_speed.tex
\begin{table}[h]
\small
\centering
\begin{adjustbox}{width=\linewidth}
\begin{tabular}{lccccc}
\toprule
\multirow{2}{*}{Metric} & Flow & 3D Points & \multicolumn{2}{c}{Action Calc.} \\
\cmidrule(lr){4-5}
&Pred. & Proj. & EEF Only & Full Joints \\
\midrule
Inf. Latency & 4.37\,s (only once) & 0.01\,s & 0.21\,s & 0.37\,s  \\
Succ. Rate (sim / real)
 & - & - & 66.7\% / 70.0\% & 72.0\% / 77.1\%  \\
\bottomrule
\end{tabular}
\end{adjustbox}
\caption{Inference latency and success rate of EC-Flow.}
\label{tab:inference_speed}
\end{table}

%% file: material/table/cross_embodiment.tex
\begin{table}[htbp]
\small
\centering
\begin{adjustbox}{width=\linewidth}
\begin{tabular}{lccccc}
\toprule

Data Comp.& \textcolor{red}{50}\faUser & \textcolor{blue}{2}\faRobot & \textcolor{red}{5}\faUser + \textcolor{blue}{2}\faRobot & \textcolor{red}{50}\faUser + \textcolor{blue}{2}\faRobot \\

\midrule
Succ. Rate & 0\% & 46\% & 52\% & 70\%\\

\bottomrule
\end{tabular}
\end{adjustbox}
\caption{Success rate for robot and human data compositions.}
\label{tab:cross_embodiment}
\end{table}

%% file: main.bbl
\begin{thebibliography}{42}
\providecommand{\natexlab}[1]{#1}
\providecommand{\url}[1]{\texttt{#1}}
\expandafter\ifx\csname urlstyle\endcsname\relax
  \providecommand{\doi}[1]{doi: #1}\else
  \providecommand{\doi}{doi: \begingroup \urlstyle{rm}\Url}\fi

\bibitem[Bharadhwaj et~al.(2024)Bharadhwaj, Mottaghi, Gupta, and
  Tulsiani]{track2act}
Homanga Bharadhwaj, Roozbeh Mottaghi, Abhinav Gupta, and Shubham Tulsiani.
\newblock Track2act: Predicting point tracks from internet videos enables
  generalizable robot manipulation.
\newblock In \emph{European Conference on Computer Vision}, 2024.

\bibitem[Black et~al.(2024{\natexlab{a}})Black, Brown, Driess, Esmail, Equi,
  Finn, Fusai, Groom, Hausman, Ichter, et~al.]{pi0}
Kevin Black, Noah Brown, Danny Driess, Adnan Esmail, Michael Equi, Chelsea
  Finn, Niccolo Fusai, Lachy Groom, Karol Hausman, Brian Ichter, et~al.
\newblock $\pi_0$: A vision-language-action flow model for general robot
  control.
\newblock \emph{arXiv preprint arXiv:2410.24164}, 2024{\natexlab{a}}.

\bibitem[Black et~al.(2024{\natexlab{b}})Black, Nakamoto, Atreya, Walke, Finn,
  Kumar, and Levine]{susie}
Kevin Black, Mitsuhiko Nakamoto, Pranav Atreya, Homer~Rich Walke, Chelsea Finn,
  Aviral Kumar, and Sergey Levine.
\newblock Zero-shot robotic manipulation with pre-trained image-editing
  diffusion models.
\newblock In \emph{The Twelfth International Conference on Learning
  Representations}, 2024{\natexlab{b}}.

\bibitem[Brohan et~al.(2022)Brohan, Brown, Carbajal, Chebotar, Dabis, Finn,
  Gopalakrishnan, Hausman, Herzog, Hsu, et~al.]{rt1}
Anthony Brohan, Noah Brown, Justice Carbajal, Yevgen Chebotar, Joseph Dabis,
  Chelsea Finn, Keerthana Gopalakrishnan, Karol Hausman, Alex Herzog, Jasmine
  Hsu, et~al.
\newblock Rt-1: Robotics transformer for real-world control at scale.
\newblock \emph{arXiv preprint arXiv:2212.06817}, 2022.

\bibitem[Cheang et~al.(2024)Cheang, Chen, Jing, Kong, Li, Li, Liu, Wu, Xu,
  Yang, et~al.]{gr2}
Chi-Lam Cheang, Guangzeng Chen, Ya Jing, Tao Kong, Hang Li, Yifeng Li, Yuxiao
  Liu, Hongtao Wu, Jiafeng Xu, Yichu Yang, et~al.
\newblock Gr-2: A generative video-language-action model with web-scale
  knowledge for robot manipulation.
\newblock \emph{arXiv preprint arXiv:2410.06158}, 2024.

\bibitem[Chi et~al.(2023)Chi, Xu, Feng, Cousineau, Du, Burchfiel, Tedrake, and
  Song]{diffusion_policy}
Cheng Chi, Zhenjia Xu, Siyuan Feng, Eric Cousineau, Yilun Du, Benjamin
  Burchfiel, Russ Tedrake, and Shuran Song.
\newblock Diffusion policy: Visuomotor policy learning via action diffusion.
\newblock \emph{The International Journal of Robotics Research}, page
  02783649241273668, 2023.

\bibitem[Du et~al.(2024{\natexlab{a}})Du, Yang, Dai, Dai, Nachum, Tenenbaum,
  Schuurmans, and Abbeel]{unipi}
Yilun Du, Sherry Yang, Bo Dai, Hanjun Dai, Ofir Nachum, Josh Tenenbaum, Dale
  Schuurmans, and Pieter Abbeel.
\newblock Learning universal policies via text-guided video generation.
\newblock \emph{Advances in Neural Information Processing Systems}, 36,
  2024{\natexlab{a}}.

\bibitem[Du et~al.(2024{\natexlab{b}})Du, Yang, Florence, Xia, Wahid, brian
  ichter, Sermanet, Yu, Abbeel, Tenenbaum, Kaelbling, Zeng, and Tompson]{vip}
Yilun Du, Sherry Yang, Pete Florence, Fei Xia, Ayzaan Wahid, brian ichter,
  Pierre Sermanet, Tianhe Yu, Pieter Abbeel, Joshua~B. Tenenbaum, Leslie~Pack
  Kaelbling, Andy Zeng, and Jonathan Tompson.
\newblock Video language planning.
\newblock In \emph{The Twelfth International Conference on Learning
  Representations}, 2024{\natexlab{b}}.

\bibitem[Fang et~al.(2020)Fang, Wang, Gou, and Lu]{graspnet}
Hao-Shu Fang, Chenxi Wang, Minghao Gou, and Cewu Lu.
\newblock Graspnet-1billion: A large-scale benchmark for general object
  grasping.
\newblock In \emph{Proceedings of the IEEE/CVF Conference on Computer Vision
  and Pattern Recognition}, pages 11444--11453, 2020.

\bibitem[Gao et~al.(2025)Gao, Zhang, Xu, Zhehao, and Shao]{flip}
Chongkai Gao, Haozhuo Zhang, Zhixuan Xu, Cai Zhehao, and Lin Shao.
\newblock {FLIP}: Flow-centric generative planning as general-purpose
  manipulation world model.
\newblock In \emph{The Thirteenth International Conference on Learning
  Representations}, 2025.

\bibitem[Goyal et~al.(2017)Goyal, Ebrahimi~Kahou, Michalski, Materzynska,
  Westphal, Kim, Haenel, Fruend, Yianilos, Mueller-Freitag, et~al.]{sth-sth-v2}
Raghav Goyal, Samira Ebrahimi~Kahou, Vincent Michalski, Joanna Materzynska,
  Susanne Westphal, Heuna Kim, Valentin Haenel, Ingo Fruend, Peter Yianilos,
  Moritz Mueller-Freitag, et~al.
\newblock The" something something" video database for learning and evaluating
  visual common sense.
\newblock In \emph{Proceedings of the IEEE International Conference on Computer
  Vision}, pages 5842--5850, 2017.

\bibitem[He et~al.(2016)He, Zhang, Ren, and Sun]{resnet}
Kaiming He, Xiangyu Zhang, Shaoqing Ren, and Jian Sun.
\newblock Deep residual learning for image recognition.
\newblock In \emph{Proceedings of the IEEE Conference on Computer Vision and
  Pattern Recognition}, pages 770--778, 2016.

\bibitem[Ho et~al.(2020)Ho, Jain, and Abbeel]{diffusion}
Jonathan Ho, Ajay Jain, and Pieter Abbeel.
\newblock Denoising diffusion probabilistic models.
\newblock \emph{Advances in neural information processing systems},
  33:\penalty0 6840--6851, 2020.

\bibitem[Karaev et~al.(2024)Karaev, Makarov, Wang, Neverova, Vedaldi, and
  Rupprecht]{cotracker}
Nikita Karaev, Iurii Makarov, Jianyuan Wang, Natalia Neverova, Andrea Vedaldi,
  and Christian Rupprecht.
\newblock Cotracker3: Simpler and better point tracking by pseudo-labelling
  real videos.
\newblock \emph{arXiv preprint arXiv:2410.11831}, 2024.

\bibitem[Kim et~al.(2025)Kim, Pertsch, Karamcheti, Xiao, Balakrishna, Nair,
  Rafailov, Foster, Sanketi, Vuong, Kollar, Burchfiel, Tedrake, Sadigh, Levine,
  Liang, and Finn]{openvla}
Moo~Jin Kim, Karl Pertsch, Siddharth Karamcheti, Ted Xiao, Ashwin Balakrishna,
  Suraj Nair, Rafael Rafailov, Ethan~P Foster, Pannag~R Sanketi, Quan Vuong,
  Thomas Kollar, Benjamin Burchfiel, Russ Tedrake, Dorsa Sadigh, Sergey Levine,
  Percy Liang, and Chelsea Finn.
\newblock Openvla: An open-source vision-language-action model.
\newblock In \emph{Proceedings of The 8th Conference on Robot Learning}, pages
  2679--2713. PMLR, 2025.

\bibitem[Ko et~al.(2024)Ko, Mao, Du, Sun, and Tenenbaum]{avdc}
Po-Chen Ko, Jiayuan Mao, Yilun Du, Shao-Hua Sun, and Joshua~B. Tenenbaum.
\newblock Learning to act from actionless videos through dense correspondences.
\newblock In \emph{The Twelfth International Conference on Learning
  Representations}, 2024.

\bibitem[Li et~al.(2025{\natexlab{a}})Li, Chen, Wu, Ma, Wu, Huang, Wang, Kong,
  and Tan]{bridgevla}
Peiyan Li, Yixiang Chen, Hongtao Wu, Xiao Ma, Xiangnan Wu, Yan Huang, Liang
  Wang, Tao Kong, and Tieniu Tan.
\newblock Bridgevla: Input-output alignment for efficient 3d manipulation
  learning with vision-language models.
\newblock \emph{arXiv preprint arXiv:2506.07961}, 2025{\natexlab{a}}.

\bibitem[Li et~al.(2025{\natexlab{b}})Li, Wu, Huang, Cheang, Wang, and
  Kong]{gr-mg}
Peiyan Li, Hongtao Wu, Yan Huang, Chilam Cheang, Liang Wang, and Tao Kong.
\newblock Gr-mg: Leveraging partially-annotated data via multi-modal
  goal-conditioned policy.
\newblock \emph{IEEE Robotics and Automation Letters}, 2025{\natexlab{b}}.

\bibitem[Liang et~al.(2024)Liang, Liu, Ozguroglu, Sudhakar, Dave, Tokmakov,
  Song, and Vondrick]{dreamitate}
Junbang Liang, Ruoshi Liu, Ege Ozguroglu, Sruthi Sudhakar, Achal Dave, Pavel
  Tokmakov, Shuran Song, and Carl Vondrick.
\newblock Dreamitate: Real-world visuomotor policy learning via video
  generation.
\newblock In \emph{8th Annual Conference on Robot Learning}, 2024.

\bibitem[Liu et~al.(2025)Liu, Wu, Li, Tan, Chen, Wang, Xu, Su, and Zhu]{rdt}
Songming Liu, Lingxuan Wu, Bangguo Li, Hengkai Tan, Huayu Chen, Zhengyi Wang,
  Ke Xu, Hang Su, and Jun Zhu.
\newblock {RDT}-1b: a diffusion foundation model for bimanual manipulation.
\newblock In \emph{The Thirteenth International Conference on Learning
  Representations}, 2025.

\bibitem[Mendonca et~al.(2023)Mendonca, Bahl, and Pathak]{structured}
Russell Mendonca, Shikhar Bahl, and Deepak Pathak.
\newblock Structured world models from human videos.
\newblock In \emph{Robotics: Science and Systems}, 2023.

\bibitem[Nair et~al.(2023)Nair, Rajeswaran, Kumar, Finn, and Gupta]{r3m}
Suraj Nair, Aravind Rajeswaran, Vikash Kumar, Chelsea Finn, and Abhinav Gupta.
\newblock R3m: A universal visual representation for robot manipulation.
\newblock In \emph{Proceedings of The 6th Conference on Robot Learning}, pages
  892--909. PMLR, 2023.

\bibitem[Ni et~al.(2024)Ni, Hao, Wu, Kou, Liu, Zheng, Wang, and
  Zhuang]{subgoal-img-gen}
Fei Ni, Jianye Hao, Shiguang Wu, Longxin Kou, Jiashun Liu, Yan Zheng, Bin Wang,
  and Yuzheng Zhuang.
\newblock Generate subgoal images before act: Unlocking the chain-of-thought
  reasoning in diffusion model for robot manipulation with multimodal prompts.
\newblock In \emph{Proceedings of the IEEE/CVF Conference on Computer Vision
  and Pattern Recognition}, pages 13991--14000, 2024.

\bibitem[{Octo Model Team} et~al.(2024){Octo Model Team}, Ghosh, Walke,
  Pertsch, Black, Mees, Dasari, Hejna, Xu, Luo, Kreiman, Tan, Sanketi, Vuong,
  Xiao, Sadigh, Finn, and Levine]{octo}
{Octo Model Team}, Dibya Ghosh, Homer Walke, Karl Pertsch, Kevin Black, Oier
  Mees, Sudeep Dasari, Joey Hejna, Charles Xu, Jianlan Luo, Tobias Kreiman,
  {You Liang} Tan, Pannag Sanketi, Quan Vuong, Ted Xiao, Dorsa Sadigh, Chelsea
  Finn, and Sergey Levine.
\newblock Octo: An open-source generalist robot policy.
\newblock In \emph{Proceedings of Robotics: Science and Systems}, Delft,
  Netherlands, 2024.

\bibitem[Qu et~al.(2025)Qu, Song, Chen, Yao, Ye, Ding, Wang, Gu, Zhao, Wang,
  et~al.]{spatial_vla}
Delin Qu, Haoming Song, Qizhi Chen, Yuanqi Yao, Xinyi Ye, Yan Ding, Zhigang
  Wang, JiaYuan Gu, Bin Zhao, Dong Wang, et~al.
\newblock Spatialvla: Exploring spatial representations for
  visual-language-action model.
\newblock \emph{arXiv preprint arXiv:2501.15830}, 2025.

\bibitem[Radford et~al.(2021)Radford, Kim, Hallacy, Ramesh, Goh, Agarwal,
  Sastry, Askell, Mishkin, Clark, et~al.]{clip}
Alec Radford, Jong~Wook Kim, Chris Hallacy, Aditya Ramesh, Gabriel Goh,
  Sandhini Agarwal, Girish Sastry, Amanda Askell, Pamela Mishkin, Jack Clark,
  et~al.
\newblock Learning transferable visual models from natural language
  supervision.
\newblock In \emph{International Conference on Machine Learning}, pages
  8748--8763. PMLR, 2021.

\bibitem[Ravi et~al.(2025)Ravi, Gabeur, Hu, Hu, Ryali, Ma, Khedr, R{\"a}dle,
  Rolland, Gustafson, Mintun, Pan, Alwala, Carion, Wu, Girshick, Dollar, and
  Feichtenhofer]{sam2}
Nikhila Ravi, Valentin Gabeur, Yuan-Ting Hu, Ronghang Hu, Chaitanya Ryali,
  Tengyu Ma, Haitham Khedr, Roman R{\"a}dle, Chloe Rolland, Laura Gustafson,
  Eric Mintun, Junting Pan, Kalyan~Vasudev Alwala, Nicolas Carion, Chao-Yuan
  Wu, Ross Girshick, Piotr Dollar, and Christoph Feichtenhofer.
\newblock {SAM} 2: Segment anything in images and videos.
\newblock In \emph{The Thirteenth International Conference on Learning
  Representations}, 2025.

\bibitem[Ren et~al.(2025)Ren, Sundaresan, Sadigh, Choudhury, and
  Bohg]{motion_track}
Juntao Ren, Priya Sundaresan, Dorsa Sadigh, Sanjiban Choudhury, and Jeannette
  Bohg.
\newblock Motion tracks: A unified representation for human-robot transfer in
  few-shot imitation learning.
\newblock \emph{arXiv preprint arXiv:2501.06994}, 2025.

\bibitem[Ren et~al.(2024{\natexlab{a}})Ren, Jiang, Liu, Zeng, Liu, Gao, Huang,
  Ma, Jiang, Chen, et~al.]{dino}
Tianhe Ren, Qing Jiang, Shilong Liu, Zhaoyang Zeng, Wenlong Liu, Han Gao,
  Hongjie Huang, Zhengyu Ma, Xiaoke Jiang, Yihao Chen, et~al.
\newblock Grounding dino 1.5: Advance the" edge" of open-set object detection.
\newblock \emph{arXiv preprint arXiv:2405.10300}, 2024{\natexlab{a}}.

\bibitem[Ren et~al.(2024{\natexlab{b}})Ren, Liu, Zeng, Lin, Li, Cao, Chen,
  Huang, Chen, Yan, et~al.]{grounded_sam}
Tianhe Ren, Shilong Liu, Ailing Zeng, Jing Lin, Kunchang Li, He Cao, Jiayu
  Chen, Xinyu Huang, Yukang Chen, Feng Yan, et~al.
\newblock Grounded sam: Assembling open-world models for diverse visual tasks.
\newblock \emph{arXiv preprint arXiv:2401.14159}, 2024{\natexlab{b}}.

\bibitem[Song et~al.(2021)Song, Meng, and Ermon]{ddim}
Jiaming Song, Chenlin Meng, and Stefano Ermon.
\newblock Denoising diffusion implicit models.
\newblock In \emph{International Conference on Learning Representations}, 2021.

\bibitem[Vuong et~al.(2023)Vuong, Levine, Walke, Pertsch, Singh, Doshi, Xu,
  Luo, Tan, Shah, Finn, Du, Kim, Khazatsky, Yang, Zhao, Goldberg, Hoque, Chen,
  Adebola, Sukhatme, Salhotra, Dass, Pinto, Cui, Haldar, Rai, Shafiullah, Zhu,
  Zhu, Nasiriany, Song, Chi, Pan, Burgard, Mees, Huang, Pathak, Bahl, Mendonca,
  Zhou, Srirama, Dasari, Lu, Fang, Fang, Christensen, Tomizuka, Zhan, Ding, Xu,
  Zhu, Tian, Lee, Sadigh, Cui, Belkhale, Sundaresan, Darrell, Malik,
  Radosavovic, Bohg, Srinivasan, Wang, Hansen, Wu, Yan, Su, Gu, Li,
  Suenderhauf, Rana, Burgess-Limerick, Ceola, Kawaharazuka, Kanazawa,
  Matsushima, Matsuo, Iwasawa, Furuta, Oh, Harada, Osa, Tang, Kroemer, Sharma,
  Zhang, Kim, Cho, Han, Kim, Lim, Johns, Palo, Stulp, Raffin, Bustamante,
  Silv{\'e}rio, Padalkar, Peters, Sch{\"o}lkopf, B{\"u}chler, Schneider, Guist,
  Wu, Tian, Shi, Li, Wang, Zhang, Amor, Zhou, Majd, Ott, Schiavi,
  Mart{\'\i}n-Mart{\'\i}n, Shah, Bisk, Bingham, Yu, Jain, Xiao, Hausman, Chan,
  Herzog, Xu, Kirmani, Vanhoucke, Julian, Lee, Ding, Chebotar, Tan, Liang,
  Mordatch, Rao, Lu, Gopalakrishnan, Welker, Joshi, Devin, Irpan, Moore, Wahid,
  Wu, Chen, Wohlhart, Bewley, Zhou, Leal, Kalashnikov, Sanketi, Fu, Xu, Xu,
  brian ichter, Hsu, Xu, Brohan, Sermanet, Heess, Ahn, Rafailov, Pooley, Byrne,
  Davchev, Oslund, Schaal, Jain, Go, Xia, Tompson, Armstrong, and
  Driess]{open_x_embodiment}
Quan Vuong, Sergey Levine, Homer~Rich Walke, Karl Pertsch, Anikait Singh, Ria
  Doshi, Charles Xu, Jianlan Luo, Liam Tan, Dhruv Shah, Chelsea Finn, Max Du,
  Moo~Jin Kim, Alexander Khazatsky, Jonathan~Heewon Yang, Tony~Z. Zhao, Ken
  Goldberg, Ryan Hoque, Lawrence~Yunliang Chen, Simeon Adebola, Gaurav~S.
  Sukhatme, Gautam Salhotra, Shivin Dass, Lerrel Pinto, Zichen~Jeff Cui,
  Siddhant Haldar, Anant Rai, Nur Muhammad~Mahi Shafiullah, Yuke Zhu, Yifeng
  Zhu, Soroush Nasiriany, Shuran Song, Cheng Chi, Chuer Pan, Wolfram Burgard,
  Oier Mees, Chenguang Huang, Deepak Pathak, Shikhar Bahl, Russell Mendonca,
  Gaoyue Zhou, Mohan~Kumar Srirama, Sudeep Dasari, Cewu Lu, Hao-Shu Fang,
  Hongjie Fang, Henrik~I Christensen, Masayoshi Tomizuka, Wei Zhan, Mingyu
  Ding, Chenfeng Xu, Xinghao Zhu, Ran Tian, Youngwoon Lee, Dorsa Sadigh, Yuchen
  Cui, Suneel Belkhale, Priya Sundaresan, Trevor Darrell, Jitendra Malik, Ilija
  Radosavovic, Jeannette Bohg, Krishnan Srinivasan, Xiaolong Wang, Nicklas
  Hansen, Yueh-Hua Wu, Ge Yan, Hao Su, Jiayuan Gu, Xuanlin Li, Niko
  Suenderhauf, Krishan Rana, Ben Burgess-Limerick, Federico Ceola, Kento
  Kawaharazuka, Naoaki Kanazawa, Tatsuya Matsushima, Yutaka Matsuo, Yusuke
  Iwasawa, Hiroki Furuta, Jihoon Oh, Tatsuya Harada, Takayuki Osa, Yujin Tang,
  Oliver Kroemer, Mohit Sharma, Kevin~Lee Zhang, Beomjoon Kim, Yoonyoung Cho,
  Junhyek Han, Jaehyung Kim, Joseph~J Lim, Edward Johns, Norman~Di Palo, Freek
  Stulp, Antonin Raffin, Samuel Bustamante, Jo{\~a}o Silv{\'e}rio, Abhishek
  Padalkar, Jan Peters, Bernhard Sch{\"o}lkopf, Dieter B{\"u}chler, Jan
  Schneider, Simon Guist, Jiajun Wu, Stephen Tian, Haochen Shi, Yunzhu Li,
  Yixuan Wang, Mingtong Zhang, Heni~Ben Amor, Yifan Zhou, Keyvan Majd, Lionel
  Ott, Giulio Schiavi, Roberto Mart{\'\i}n-Mart{\'\i}n, Rutav Shah, Yonatan
  Bisk, Jeffrey~T Bingham, Tianhe Yu, Vidhi Jain, Ted Xiao, Karol Hausman,
  Christine Chan, Alexander Herzog, Zhuo Xu, Sean Kirmani, Vincent Vanhoucke,
  Ryan Julian, Lisa Lee, Tianli Ding, Yevgen Chebotar, Jie Tan, Jacky Liang,
  Igor Mordatch, Kanishka Rao, Yao Lu, Keerthana Gopalakrishnan, Stefan Welker,
  Nikhil~J Joshi, Coline~Manon Devin, Alex Irpan, Sherry Moore, Ayzaan Wahid,
  Jialin Wu, Xi Chen, Paul Wohlhart, Alex Bewley, Wenxuan Zhou, Isabel Leal,
  Dmitry Kalashnikov, Pannag~R Sanketi, Chuyuan Fu, Ying Xu, Sichun Xu, brian
  ichter, Jasmine Hsu, Peng Xu, Anthony Brohan, Pierre Sermanet, Nicolas Heess,
  Michael Ahn, Rafael Rafailov, Acorn Pooley, Kendra Byrne, Todor Davchev,
  Kenneth Oslund, Stefan Schaal, Ajinkya Jain, Keegan Go, Fei Xia, Jonathan
  Tompson, Travis Armstrong, and Danny Driess.
\newblock Open x-embodiment: Robotic learning datasets and {RT}-x models.
\newblock In \emph{Towards Generalist Robots: Learning Paradigms for Scalable
  Skill Acquisition @ CoRL2023}, 2023.

\bibitem[Wen et~al.(2023)Wen, Lin, So, Chen, Dou, Gao, and Abbeel]{tap}
Chuan Wen, Xingyu Lin, John So, Kai Chen, Qi Dou, Yang Gao, and Pieter Abbeel.
\newblock Any-point trajectory modeling for policy learning.
\newblock \emph{arXiv preprint arXiv:2401.00025}, 2023.

\bibitem[Wu et~al.(2024)Wu, Jing, Cheang, Chen, Xu, Li, Liu, Li, and Kong]{gr1}
Hongtao Wu, Ya Jing, Chilam Cheang, Guangzeng Chen, Jiafeng Xu, Xinghang Li,
  Minghuan Liu, Hang Li, and Tao Kong.
\newblock Unleashing large-scale video generative pre-training for visual robot
  manipulation.
\newblock In \emph{International Conference on Learning Representations}, 2024.

\bibitem[Xu et~al.(2024)Xu, Xu, Xu, Chi, Wetzstein, Veloso, and
  Song]{im2flow2act}
Mengda Xu, Zhenjia Xu, Yinghao Xu, Cheng Chi, Gordon Wetzstein, Manuela Veloso,
  and Shuran Song.
\newblock Flow as the cross-domain manipulation interface.
\newblock In \emph{8th Annual Conference on Robot Learning}, 2024.

\bibitem[Yang et~al.(2024)Yang, Du, Ghasemipour, Tompson, Kaelbling,
  Schuurmans, and Abbeel]{unisim}
Sherry Yang, Yilun Du, Seyed Kamyar~Seyed Ghasemipour, Jonathan Tompson,
  Leslie~Pack Kaelbling, Dale Schuurmans, and Pieter Abbeel.
\newblock Learning interactive real-world simulators.
\newblock In \emph{The Twelfth International Conference on Learning
  Representations}, 2024.

\bibitem[Yu et~al.(2019)Yu, Quillen, He, Julian, Hausman, Finn, and
  Levine]{metaworld}
Tianhe Yu, Deirdre Quillen, Zhanpeng He, Ryan Julian, Karol Hausman, Chelsea
  Finn, and Sergey Levine.
\newblock Meta-world: A benchmark and evaluation for multi-task and meta
  reinforcement learning.
\newblock In \emph{Conference on Robot Learning}, 2019.

\bibitem[Yu et~al.(2024)Yu, Zhang, Liu, Li, Geng, Wang, Ding, and
  Chen]{dexgrasp}
XinQiang Yu, Jialiang Zhang, Haoran Liu, Danshi Li, Haoran Geng, He Wang, Yufei
  Ding, and Jiayi Chen.
\newblock Dexgraspnet 2.0: Learning generative dexterous grasping in
  large-scale synthetic cluttered scenes.
\newblock In \emph{2nd Workshop on Dexterous Manipulation: Design, Perception
  and Control (RSS)}, 2024.

\bibitem[Yuan et~al.(2024)Yuan, Wen, Zhang, and Gao]{general_flow}
Chengbo Yuan, Chuan Wen, Tong Zhang, and Yang Gao.
\newblock General flow as foundation affordance for scalable robot learning.
\newblock In \emph{8th Annual Conference on Robot Learning}, 2024.

\bibitem[Zhao et~al.(2024)Zhao, Chen, Meng, Mao, Song, and Zhang]{vlmpc}
Wentao Zhao, Jiaming Chen, Ziyu Meng, Donghui Mao, Ran Song, and Wei Zhang.
\newblock Vlmpc: Vision-language model predictive control for robotic
  manipulation.
\newblock In \emph{Robotics: Science and Systems}, 2024.

\bibitem[Zhu et~al.(2022)Zhu, Joshi, Stone, and Zhu]{deoxys}
Yifeng Zhu, Abhishek Joshi, Peter Stone, and Yuke Zhu.
\newblock {VIOLA}: Object-centric imitation learning for vision-based robot
  manipulation.
\newblock In \emph{6th Annual Conference on Robot Learning}, 2022.

\bibitem[Zitkovich et~al.(2023)Zitkovich, Yu, Xu, Xu, Xiao, Xia, Wu, Wohlhart,
  Welker, Wahid, et~al.]{rt2}
Brianna Zitkovich, Tianhe Yu, Sichun Xu, Peng Xu, Ted Xiao, Fei Xia, Jialin Wu,
  Paul Wohlhart, Stefan Welker, Ayzaan Wahid, et~al.
\newblock Rt-2: Vision-language-action models transfer web knowledge to robotic
  control.
\newblock In \emph{Conference on Robot Learning}, pages 2165--2183. PMLR, 2023.

\end{thebibliography}
